\documentclass[review,3p,times]{elsarticle}

\usepackage{amssymb}
\usepackage{latexsym}
\usepackage{amsmath, graphicx}
\usepackage{multirow}
\usepackage[normalem]{ulem}
\useunder{\uline}{\ul}{}
\usepackage[T1]{fontenc}
\usepackage{url}
\usepackage{xcolor}
\usepackage{bbding}

\journal{Pattern Recognition}

\begin{document}

\begin{frontmatter}
\title{Efficient Human-Object-Interaction (EHOI) Detection via Interaction Label Coding and Conditional Decision}

\author[1]{Tsung-Shan Yang\corref{cor1}} 
\cortext[cor1]{Corresponding author: 
  Tel.: +1-213-519-1489;}
\ead{tsungsha@usc.edu}

\author[1]{Yun-Cheng Wang}
\author[1]{Chengwei Wei}
\author[2]{Suya You}
\author[1]{C.-C. Jay Kuo}

\affiliation[1]{organization={University of Southern California},
                city={Los Angeles}, 
                postcode={90089}, 
                state={CA},
                country={USA}}

\affiliation[2]{organization={DEVCOM Army Research Laboratory},
                city={Adelphi}, 
                postcode={20783}, 
                state={Maryland},
                country={USA}}

\begin{abstract}
Human-Object Interaction (HOI) detection is a fundamental task in image understanding. While deep-learning-based HOI methods provide high performance in terms of mean Average Precision (mAP), they are computationally expensive and opaque in training and inference processes. An Efficient HOI (EHOI) detector is proposed in this work to strike a good balance between detection performance, inference complexity, and mathematical transparency. EHOI is a two-stage method. In the first stage, it leverages a frozen object detector to localize the objects and extract various features as intermediate outputs. In the second stage, the first-stage outputs predict the interaction type using the XGBoost classifier. Our contributions include the application of error correction codes (ECCs) to encode rare interaction cases, which reduces the model size and the complexity of the XGBoost classifier in the second stage.
Additionally, we provide a mathematical formulation of the relabeling and decision-making process. Apart from the architecture, we present qualitative results to explain the functionalities of the feedforward modules. Experimental results demonstrate the advantages of ECC-coded interaction labels and the excellent balance of detection performance and complexity of the proposed EHOI method. 
\end{abstract}


\end{frontmatter}

\section{Introduction}\label{sec:intro}

\begin{figure}[htbp]
\centering
\includegraphics[width=\columnwidth]{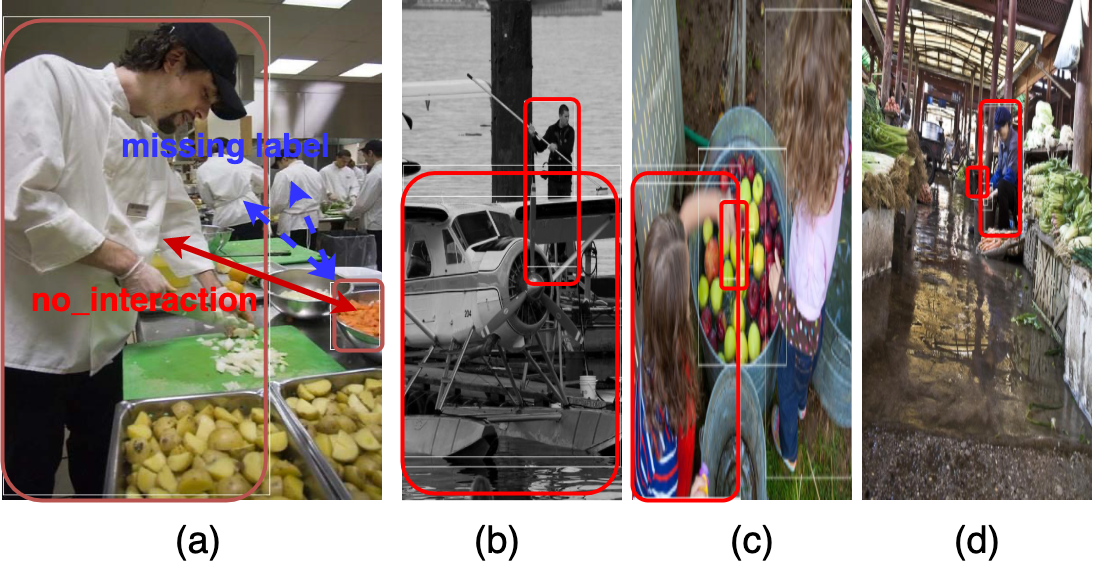}
\caption{Illustration of challenges in the HOI problem with images from the HICO-DET dataset: (a) images labeled as `no\_interaction,' (b)-(d) three images with the same verb, `wash,' but humans behave differently.}
\label{fig: examples}
\end{figure}

Human-Object Interaction (HOI) detection is essential for image understanding~\citep{chao2018learning, gupta2019no}. The labels in HOI datasets are triplets in the form of <Human-Interaction-Object>. The detector needs to know not only the bounding boxes of the human and objects but also the class labels of the objects and the types of interaction. HOI detection is a human-centric application and can be further applied to the image understanding task, such as visual question answering and image captioning. HOI detection can be challenging, as illustrated in Figure\ref{fig: examples}. The examples are taken from a popular HOI detection dataset, HICO-DET~\citep{chao2018learning}. Images may contain the label \textit{`no\_interaction'}, but not every \textit{`no\_interaction'} human-object pair is labeled. Furthermore, some images may share the same verb in different scenarios, known as verb polysemy~\citep{zhong2021polysemy}.

\begin{figure}[htbp]
\centering
\includegraphics[width=\columnwidth]{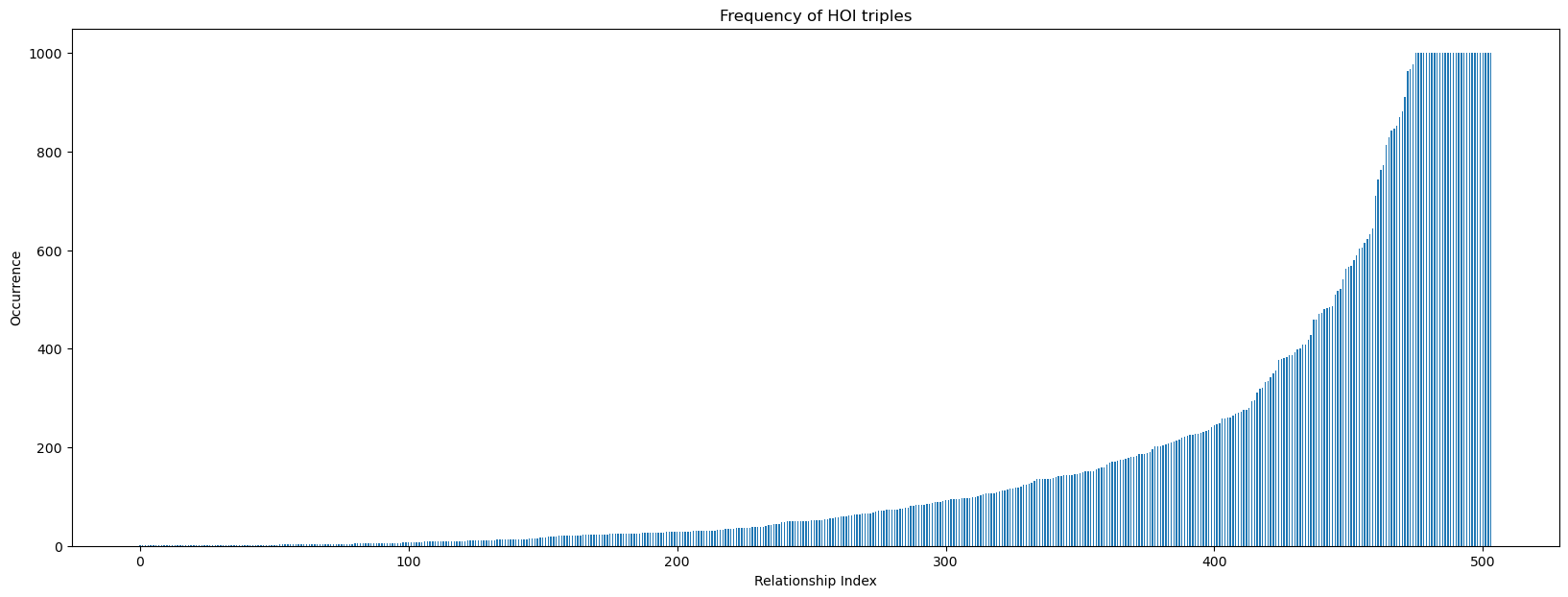}
\caption{The occurrence of <human-interaction-object> labels, where if the number of annotations exceeds 1,000, it is clipped to 1,000 for simplicity. Half of the relationship labels in the HICO-DET dataset have fewer than 200 annotations.}\label{fig: distribution}
\end{figure}

Another challenge in HOI detection is the imbalanced distribution of interaction pairs in the training samples. As shown in Figure~\ref{fig: distribution}, the HICO-DET dataset exhibits a skewed distribution of annotations, leading to highly biased predictions. To prevent overfitting on the relationship annotations, we propose a hybrid coding scheme to address this problem. That is, we partition interaction pairs into rare and non-rare cases. For non-rare cases, we adopt the traditional one-hot coding. For rare cases, we group them into one super-class and then adopt binary codes and error correction codes (ECCs) to encode these rare cases. The repartitioning and data relabeling mitigate the imbalanced distribution in the original dataset. This is one of the significant contributions of this work. 

\begin{figure}[htbp]
\centering
\includegraphics[width=0.7\columnwidth]{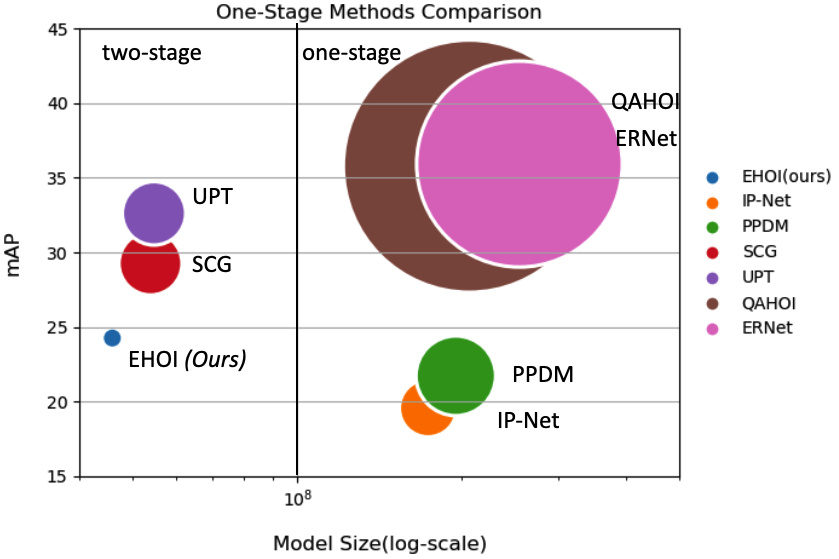}
\caption{Complexity comparison between the proposed EHOI and several other state-of-the-art (SOTA) detectors for the HICO-DET dataset, where the x-axis is the model size in the log scale, the y-axis is mAP (\%), and the bubble size is proportional to the inference FLOP numbers.}
\label{fig: complexity}
\end{figure}

Prevailing HOI detection solutions can be categorized into one-stage and two-stage methods. One-stage methods~\citep{lim2023ernet, chen2023qahoi, ma2023fgahoi, kim2021hotr} are trained via end-to-end optimization of specific neural network architectures, where the objective function relates to the <human-interaction-object> labels. The models fit the training dataset. Yet, labels in the dataset are not aligned perfectly with human understanding. An imperfect label is shown in Figure~\ref{fig: examples}, which indicates that the algorithms may be biased due to the imbalanced labels, making them difficult to interpret. Two-stage methods~\citep{zhang2022efficient, zhang2021spatially, hou2021detecting} include object detection and relationship prediction steps, namely, 1) identifying where humans and objects are located and 2) determining the type of interaction between them. In the first stage, it conducts object detection using a pre-trained object detector and extracts various features from input images. In the second stage, it leverages the first-stage outputs, such as object classes, bounding boxes, spatial relationships, etc., to predict the interaction type. Deep Learning (DL) models are widely applied in both stages. However, when deployed on edge applications, the cascaded model structures are computationally expensive. Typically, one-stage models outperform two-stage models in detection accuracy at the expense of larger model sizes and higher training/inference complexities. Figure~\ref{fig: complexity} visualizes the model complexity and inference cost. On the other hand, the decoupled modular design makes two-stage models more straightforward to understand.

Figure ~\ref{fig: complexity} compares the model performance, model size, and computational complexity of several state-of-the-art (SOTA) HOI detectors. The model performance is represented by mAP (\%) in the y-axis, the model size is represented by the number of model parameters in the x-axis, and the computational complexity is represented by inference floating-point operation (FLOP) numbers, respectively. This figure shows one-stage transformer-based models (QAHOI~\citep{chen2023qahoi} and ERNet~\citep{lim2023ernet}) and one-stage models based on interaction point prediction (IP-Net~\citep{wang2020learning} and PPDM~\citep{liao2020ppdm}).  We also visualize two SOTA two-stage methods - UPT~\citep{zhang2022efficient} and SCG~\citep{zhang2021spatially}. 

Aiming at interpretability and lower carbon footprints, we propose a new two-stage method called Efficient HOI (EHOI) in this work. As shown in Figure~\ref{fig: complexity}, its mAP performance is worse than other two-stage models, UPT and SCG, yet its FLOP number is significantly lower. As reported in Sec. \ref{sec:exp}, the FLOP number of EHOI is 4,500 times smaller than SCG and 15,800 times smaller than UPT per query. EHOI outperforms some one-stage models (IP-Net and PPDM) and underperforms others (QAHOI and ERNet) in mAP. However, it has tremendous advantages in the model size and FLOP numbers. Regarding carbon footprint (FLOP number) and memory (model size), EHOI offers an attractive AI/ML solution for mobile and edge devices. EHOI extends green learning~\citep{kuo2023green} to the HOI detection problem. The detector aims to provide a transparent decision-making process and maintain low power consumption regarding the number of floating point operations. 

The contributions of this work are summarized below.
\begin{itemize}
    \item EHOI offers comparable mAP performance with significantly lower inference complexity for the HICO-DET dataset.
    \item EHOI is explainable. All of its modules can be interpreted as probability estimators. We formulate the learning process as the aggregation of conditional probabilities.
    \item The data distribution in the HOI dataset is analyzed. We shed light on individual modules with visual examples and explain modules' decoupling statistically. 
\end{itemize}

\section{Related Work}\label{sec:relat}

\subsection{One-stage HOI Detection}

One-stage HOI detectors are inspired by one-stage object detectors~\citep{ren2015faster,carion2020end}. Apart from the human and object bounding boxes, researchers define the interaction vectors from the center of the human bounding box to the corresponding object bounding box. Starting from the detector backbones, deep learning models can perform HOI detection with auxiliary interaction points or vectors. Wang et al.~\citep{wang2020learning} cropped the intersection of the proposed human and object bounding boxes as the interaction region. The model can predict the internship label by exploiting the overlapped features. Liao et al.~\citep{liao2020ppdm} proposed a model optimized by the auxiliary output of predicting the interaction points. Leveraging the spatial information, the supervised loss from interaction points helps the model obtain better features and remove unlikely interactions.

With the thriving visual transformers (ViTs), transformer-based detectors~\citep{zhu2020deformable, jia2023detrs} have been developed. The encoder-decoder structure conducts the detection by trainable query tensors. To merge the information from human and object bounding boxes, Kim et al.~\citep{kim2021hotr} developed a DETR~\citep{carion2020end} backbone and combined it with pairwise human/object queries for the interaction decoder. The merge human and object queries are interaction queries for further decoding. Chen et al.~\citep{chen2021reformulating} introduced auxiliary interaction vector prediction to optimize the interaction decoder. Tamura et al.~\citep{tamura2021qpic} used the Hungarian algorithm~\citep{kuhn1955hungarian} to calculate the loss of matched human/object pairs in the loss function. By exploiting overlapped patches between humans and objects, the auxiliary output can utilize human priors' supervision by minimizing the loss function.

Besides the decoding process, utilizing the detectors in conjunction with pre-trained models significantly enhances performance. Liao et al.~\citep{liao2022gen} combined the SOTA language and visual transformer, CLIP~\citep{radford2021learning}. Matching the textual embeddings and the interaction representations improves the relation decision significantly. Inspired by the deformable operations~\citep{dai2017deformable, xia2022vision}, Chen et al.~\citep{chen2023qahoi} replaced the ResNet backbone with deformable transformers. The backbone can focus on the objects not limited to the square grids. To enrich the presentations, Lim et al.~\citep{lim2023ernet} used the EfficientNet as the backbone to extract multi-scale features. 

Although one-stage models offer SOTA HOI detection performance, they have shortcomings. First, the training of one-stage models is highly dependent on the dataset. Zhu et al. \citep{zhu2023diagnosing} pointed out that one-stage models could be biased in detection results under a skewed data distribution. Second, it is challenging to interpret one-stage methods since the semantic information in images is hidden in numerous cascading latent spaces. Researchers indirectly analyzed the models using convolutional filter responses and attention matrices in Visual Transformers (ViTs). Third, they suffer from a large model size and high computational complexity. 

\subsection{Two-stage HOI Detection}

HOI is a human-centric classification task. The linkage between human and object representations is crucial. With the stagewise algorithm, humans can understand the decision-making process clearly. In two-stage models, the first stage extracts various human and object representations. In contrast, the second stage is a multi-pair (i.e., human-object pairs) and multi-label (i.e., interaction labels) classification problem.

To understand the human representations, Gupta et al.~\citep{gupta2019no} used pose estimation models to obtain semantic information. The addition pose-estimation model provides a three-dimensional spatial understanding. By exploiting the human-object correlation, Hou et al.~\citep{hou2020visual} constructed a model to yield human and object streams and handled the relation between the two streams based on the co-occurrence of <human-relation-object> triplets. The triplets formation can be optimized by the composition learning~\citep{kato2018compositional}. The merging process removes the impossible triplets and enhances the probabilities of possible triplets. 

The human-object relationship can also be formulated as a graph, where human and object features can be viewed as the vertices in the graph. Hence, the interaction detection can be formulated as an edge production problem. Gao et al.~\citep{gao2020drg} proposed a dual structure to model the relation. It combined the human-centric and object-centric graphs to predict the relation. The relationship is the weighted sum of the two sub-graphs. Zhang et al.~\citep{zhang2021spatially} incorporated the coordinate information between objects in the graph convolution structure to capture spatial information within images. The spatial information can be defined as the vector between the human and object bounding boxes and their intersection. With the development of transformers, Zhang et al.~\citep{zhang2022efficient} adopted the FFN decoder structure for the pairwise relation classification. The decoder takes the features of the region of interest from the first stage as the inputs and performs the classification.

\subsection{Green Learning}

Kuo et al.~\citep{kuo2023green} proposed a statistical-based learning framework called Green Learning (GL) to address the increasing computational burdens of DL networks. The GL paradigm does not have neurons, neural networks, and end-to-end optimization via backpropagation. Instead, it adopts a feedforward and modular design in both training and inference based on data statistics. The whole processing pipeline is purely data-driven and transparent. The GL solution addresses environmental concerns by reducing FLOPs to relieve power consumption and carbon footprint. Our work follows this principle, as detailed in the next section. 

\section{EHOI Method}\label{sec:method}

\subsection{System Overview}

\begin{figure}[htbp]
\centering
\includegraphics[width=\linewidth]{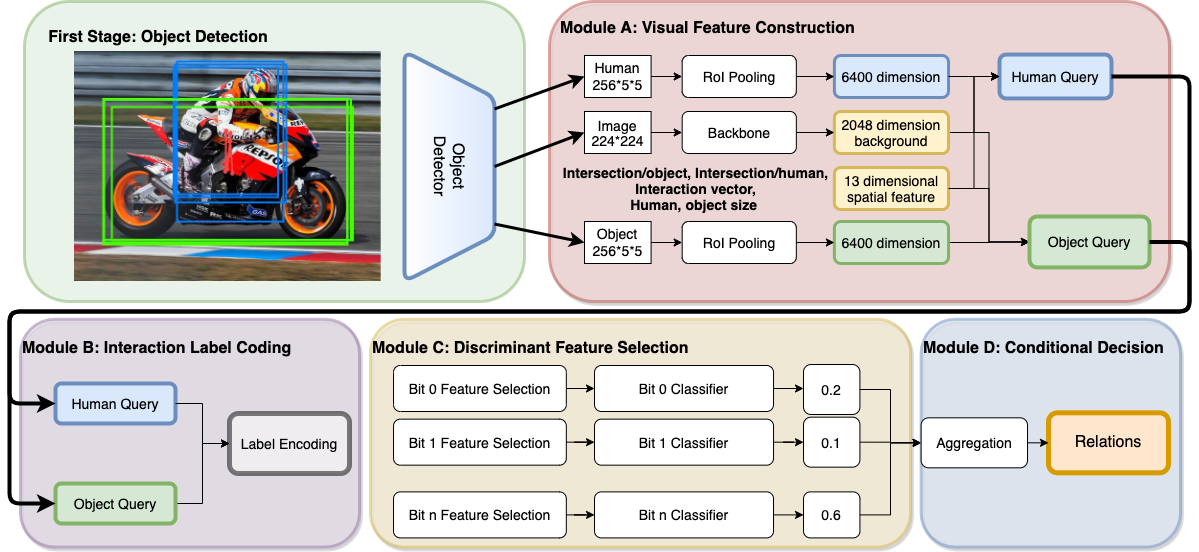}
\caption{The overall system diagram of the proposed EHOI. Its first stage is a pre-trained object detector. The main contributions of EHOI lie in the data processing pipeline in the second stage. It consists of four modules: A) visual features construction, B) interaction label coding, C) discriminant features selection, and D) conditional decision on the interaction type.} \label{fig: arch}
\end{figure}

The system diagram of the proposed EHOI method is shown in Figure~\ref{fig: arch}, which is a two-stage method. The first stage is a pre-trained object detector, where we select DETR \citep{carion2020end} as the object detector. It uses ResNet50 as the backbone and achieves good object detection performance trained by suitable object detection datasets. Since our main contributions lie in the second stage, we will emphasize the data processing pipeline of the second stage in this section. It consists of the following four tasks in cascade. 

\begin{itemize}
\item Module A: Visual Features Construction \\
We utilize the Region of Interest (RoI) alignment and pooling \citep{girshick2015fast} to generate human and object representations. It yields an input query pair that contains human and object features and their associated spatial information. 
\item Module B: Hybrid Interaction Coding \\
We propose a hybrid coding scheme to address the training samples' imbalanced distribution of interaction pairs. That is, we partition interaction pairs into rare and non-rare cases. For non-rare cases, we adopt the traditional one-hot coding. For rare cases, we group them into one super-class and then adopt binary codes with error correction codes (ECCs) to encode rare cases within the super-class. 
\item Module C: Discriminant Features Selection \\
The discriminant feature selection process is conducted based on the interaction codes. We identify a subset of discriminant features against every bit assignment of the interaction type.  
\item Module D: Conditional Decision on the Interaction Type \\
The final prediction is the aggregation of the probabilities outputs from each interaction bit. 
\end{itemize}

The model in the second stage is efficient regarding the number of model parameters and Floating-point Operations (FLOP) numbers. Modules B-C are statistics-based, allowing interpretability. Furthermore, applying error correction codes (ECCs) to encode interaction labels is a novel contribution to the AI/ML literature. Its advantages are demonstrated in the experiments section. 

\subsection{Processing Modules in the Second Stage}

\subsubsection{Module A: Visual Features Construction}
Constructing a rich feature set for human and object representations is critical. Intuitively, the relative distances and other scenarios between the human and object locations in images are helpful for HOI detection. Utilizing the first-stage model, we can capture the features of corresponding regions by RoI pooling and alignment\citep{girshick2015fast}. The human and object features can be obtained from aggregating different layer feature maps in the detector. The relative spatial information includes the interaction vector and the relative sizes of human and object bounding boxes. The interaction vector is the difference between human- and object-bounding box centers. It can be represented in Euclidean or polar coordinate systems.
Furthermore, the background information can be extracted from the whole image features obtained from the backbone network. To summarize, the features under consideration comprise human RoI features, object RoI features, relative spatial features, and whole image features. We also split human and object representations as individual queries and determine their spatial features accordingly. Some detailed descriptions are illustrated in Figure~\ref{fig: arch}. Yet, we should point out that these features are far from perfect since they lack precise semantic information, and discriminant features may be concealed in noisy training samples. 

To deal with imbalanced labels, our model fits subsets of the interaction samples instead of the whole dataset. The classifiers are trained by subsets containing a common object for human queries. That is, the desired outputs of a classifier can be denoted as
\begin{equation}
\begin{aligned}
P(relation\mid human) = \sum_{c\in\{Object\}}P(relation, object=c\mid human),
\end{aligned}
\end{equation}
where $c$ denotes an object type and $\{Object\}$ denotes the whole object set. Under the constraint, $object=c$, we can reduce the overall long-tail distribution to a few short-tail conditional distributions. Similarly, the classifier's performance for object query can also be improved by conditional probabilities. We employ clustering algorithms such as KMeans to create subsets and use them to train classifiers in each subset. We can assign a pseudo-label for each subset, and then the classifier can be formulated similarly. The ultimate classifier for object queries can be obtained by combining multiple subset classifiers in a weighted manner. We can aggregate the results of both human and object queries to make relation decisions in the inference stage. 

\subsubsection{Module B: Hybrid Interaction Coding} The foundation of modern machine learning models is to capture the distribution in the training dataset and generalize it to unseen samples. Finding a representation space that generalizes well between training and testing samples is essential. Representations could be highly diversified, and the labeled data may possess a long-tail distribution in real-world applications. Adopting the one-hot vector to represent the classes of interest in the context of AI/ML is typical. There are two problems with the one-hot representation. First, if the class number is large, the dimension of these one-hot vectors can be high. Second, the labeled data possess a long-tail distribution, as mentioned above. Take the HOI benchmark, HICO-DET, as an example. It has 600 interaction triplets. However, 138 have less than ten samples and are called rare cases. If we adopt the one-hot encoding scheme for all, we need 600-dimensional vectors to represent them and have to train 600 one-versue-the-rest binary classifiers. The classification performance of each rare case is expected to be poor due to high data imbalance since it is challenging for a classifier to learn from less than ten samples among more than $10^6$ queries. 

To handle this challenge, we merge all rare cases into a super-class and adopt the traditional one-hot coding to encode non-rare cases plus this super-class. Then, we adopt the binary coding scheme to differentiate rare cases inside the super-class. To compare the difference between the one-hot and binary coding, we take a 4-class classification problem as an example. The four classes are $\{1000, 0100, 0010, 0001\}$ in the one-hot encoding and as $\{00, 01, 10, 11\}$ in the binary coding. Each bit represents a binary split. We can train four binary classifiers for the one-hot coding that handles the one-versus-the-rest classification problem. For the binary coding, we only train two binary classifiers. The first one separates $\{00, 01\}$ from $\{10, 11\}$ based on the first bit while the second one splits $\{00, 10\}$ from $\{01, 11\}$ based on the second bit.  Each binary classifier can be viewed as a decoder. Nevertheless, each classifier may have mistakes, leading to wrong aggregated results. We use error correction codes (ECC) to enhance the robustness of a remedy. To follow the above example, we can assign three-bit codewords to them, i.e., $\{000, 011, 101, 110\}$. After adding the error correction bit, every codeword pair has a Hamming distance of 2 (i.e., have two different bits). Here, we use Hamming codes~\citep{hamming1950error} to improve the performance of straightforward binary codes and ensure that each representation differs from others with a Hamming distance of no less than 3 in EHOI. 

The performance of four coding schemes is compared in Table~\ref{table: coding} for the HICO-DET dataset. They are one-hot codes, binary codes, Hamming codes, and a hybrid coding scheme. The hybrid coding scheme adopts the one-hot coding for non-rare cases, the super-class of all rare cases, and the Hamming codes for rare cases. The table shows that the hybrid coding scheme achieves the best results, with a higher mAP value than one-hot codes. It is also worthwhile to point out that Hamming codes have a smaller model size, which helps reduce the model size of the hybrid scheme. 

\begin{table}[htbp]\label{table: coding}
\caption{Performance comparison between four coding schemes for interaction labels in mAP (\%). Under the same architecture, the Hamming codes perform best for rare cases, while the one-hot codes offer the best performance for non-rare cases. The hybrid coding schemes yield the best overall performance.}
\centering
\begin{tabular}{llllllll}
\multirow{2}{*}{Methods}   & \multicolumn{3}{c}{Default}   &\multirow{2}{*}{Model Size}\\ \cline{2-4}
                           & Full           & Rare           & Non-Rare       &        \\ \hline
EHOI (one-hot codes)       & 20.55          & 13.47          & 22.66          & 56.4M  \\
EHOI (binary codes)        & 16.35          & 7.97           & 18.86          & 15.4M  \\
EHOI (Hamming codes)       & 19.19          & 14.50          & 20.59          & 27.7M  \\ \hline
EHOI (hybrid)              & \textbf{24.53} & \textbf{19.26} & \textbf{26.09} & 45.9M
\end{tabular}
\end{table}

\subsubsection{Module C: Discriminant Features Selection} Each bit representation in Hamming codes corresponds to a partition of labeled interactions into two sets. In other words, we relabel interactions of rare cases into two types denoted by 0 and 1, respectively. If we don't change the features, this relabeling will become a subset of the original coding scheme. Hence, each bit assignment prompts the utilization of a feature selection module to identify the pivotal feature and exclude others. Such dimension reduction can prevent overfitting the estimator with the low-variance features.

\begin{figure}[htbp]
\centering
\includegraphics[width=\columnwidth]{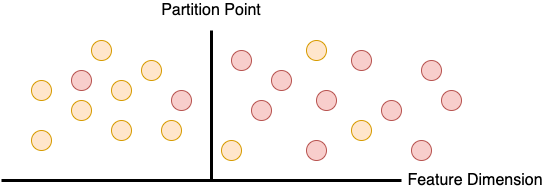}
\caption{Visualization of DFT, where pink and orange dots represent the ``0'' and ``1'' binary labels, and the loss function is the weighted cross-entropy sum of samples in the left and right parts of the partition line.}\label{fig: DFT}
\end{figure}

We must select discriminant features for a given binary label to facilitate the classifier in the next module. This can be achieved by applying the Discriminant Feature Test (DFT)~\citep{yang2022supervised} to all input features individually. For a given 1D input feature, we place the feature value of each labeled training sample in a line segment bounded by the range of the maximum and minimum values, as shown in Figure  \ref{fig: DFT}. Then, we search for the optimal partition point on this line segment to minimize the loss function, defined as the weighted sum of binary cross-entropies of the left and right partitions.
\begin{equation}
    H = -\sum_{i\in\{\text{partition}\}} y_{i}\log(y_{i}),
\end{equation}
where $H$ denotes the binary entropy of the partitioned samples and $y$ represents the samples' labels.
\begin{equation}
    L = \frac{N_{left}H_{left}+N_{right}H_{right}}{N_{left}+N_{right}},
\end{equation} where $L$ denotes the loss of a partition point, and $N_{left}, N_{right}, H_{left}, H_{right}$ denote the number of the samples and the binary cross-entropy in the left and right partition, respectively. A feature is more discriminant if it has a lower loss value, reducing the number of miss-classified cases. Then, we can plot the loss value curve from the lowest to the highest and use the elbow point to select discriminant features from the whole feature set. 

However, the detection dataset accompanies plenty of false detections. The imbalance distribution also occurs in the HOI detection. To enhance the correlation between features and positive labels, we change the binary cross-entropy to focal loss~\citep{lin2017focal}. 
\begin{equation}
    H^{f} = -\alpha (1-p)^{\gamma}\log(p),
\end{equation}
where $H^{f}$ denotes the focal loss in the partition, and $\alpha , \gamma \in\mathbf{R}$ are hyperparameters. The exponential parameter can emphasize the hard positive cases. Accordingly, the loss of the partition points can be denoted as the weighted sums of the two focal losses in the left and right partition.
\begin{equation}
    L^{f} = \frac{N_{left}H^{f}_{left}+N_{right}H^{f}_{right}}{N_{left}+N_{right}}.
\end{equation}

\subsubsection{Module D: Conditional Decision on the Interaction Type}\label{sec: decompose} 
We divide the desired decisions into sequential subproblems instead of training a complex classifier for the skewed data distribution. Each subproblem can be expressed clearly, step by step. First, we attempt to maximize the conditional probability of an interaction (or relation) conditioned on the human and object representations, which can be written as
\begin{equation}\label{eq: decouple}
    \begin{aligned}
        &P(relation \mid human, object) \\
        &=P(relation \mid human)*\frac{P(object\mid relation, human)}{P(object \mid human)}\\
        &=P(relation \mid object)*\frac{P(human \mid relation, object)}{P(human \mid object)}\\
        &\sim \alpha P(relation \mid human) + \beta P(relation \mid object),
    \end{aligned}
\end{equation}
where $\alpha, \beta$ are learnable parameters. Then, the two conditional probabilities in the last equation can be further expressed as
\begin{equation}
    \begin{aligned}
    P(relation|human) &= \sum_{c\in\{Object\}}P(relation, object=c|human)\\
    P(relation|object) &= \sum_{d\in\{Human\}}P(relation, human=d|object), 
    \end{aligned}
\end{equation}
where $c$ and $d$ are class labels and $\{Object\}, \{Human\}$ are the object sets and clustered human representations. Suppose we use a bit stream, $B=(b_0, b_1, \cdots, b_{n-1})$, to represent a relation.  Then, we would like to maximize $P(relation, object=c|human)$ and $P(relation, human=d|object)$, which are called the human query and the object query, respectively. The conditional probability of human queries can be written as
\begin{equation}
    \begin{aligned}
        & P(relation, object=c|human)\\ 
        & = P(B, object=c|human)\\
        & = \bigcap\limits_{0 \leq i < n}P(b_i, object=c|human).
    \end{aligned}
\end{equation}
The conditional probability of object queries can be found in the same manner. 

All the probability estimators in EHOI are XGBoost~\citep{chen2016xgboost}. For each bit classifier, we set the number of estimators and the depth of the tree to 300 and 3, respectively. The aggregation of the bit stream prediction is conducted by Linear Discriminant Analysis (LDA). 

\begin{table}[htbp]
\caption{Detection performance comparison of SOTA two-stage models in mAP (\%) for the HICO-DET dataset under the default and known object settings and for the V-COCO dataset. The model sizes in the parameter number (M) are also compared, where the numbers are taken from Lim et al. \citep{lim2023ernet}. }\label{table: Two-stage}
\resizebox{\linewidth}{!}{
\begin{tabular}{cccccccccccc}
\multirow{2}{*}{Architecture}       & \multirow{2}{*}{Method} & \multirow{2}{*}{Param(M)($\downarrow$)} & \multirow{2}{*}{Backbone} & \multicolumn{3}{c}{Default($\uparrow$)}                      & \multicolumn{3}{c}{Known Object($\uparrow$)}                 & \multicolumn{2}{c}{V-COCO($\uparrow$)}                                                  \\ \cline{5-12} 
                                    &                         &                         &                           & Full           & Rare           & Non-Rare       & Full           & Rare           & Non-Rare       & $AP^{S1}_{role}$ & $AP^{S2}_{role}$ \\ \hline
\multicolumn{12}{c}{\textbf{Two-Stage Methods}}                                                                                                                                                                                                                                        \\ \hline
\multirow{5}{*}{Multi-Stream}       & No-Frill\citep{gupta2019no}        &72.3 & ResNet152                 & 17.18          & 12.17          & 18.08          & -              & -              & -              & -                                    & -                                    \\
                                    & PMFNet\citep{wan2019pose}          &49.3 & ResNet50                  & 17.46          & 15.65          & 18.00          & 20.34          & 17.47          & 21.20          & -                                    & -                                    \\
                                    & ACP\citep{bansal2020detecting}     &-    & ResNet101                 & 21.96          & 16.43          & 23.62          & -              & -              & -              & 53.2                                 & -                                    \\
                                    & PD-Net\citep{zhong2021polysemy}    & -   & ResNet152                 & 22.37          & 17.61          & 23.79          & 26.86          & 21.70          & 28.44          & 52.0                                 & -                                    \\
                                    & VCL\citep{hou2020visual}           & -   & ResNet50                  & 23.63          & 17.21          & 25.55          & 25.98          & 19.12          & 28.03          & 48.3                                 & -                                    \\ \cline{2-12} 
\multirow{4}{*}{Graph-Based}        & RPNN\citep{zhou2019relation}       & -   & ResNet-50                 & 17.35          & 12.78          & 18.71          & -              & -              & -              &                                      &                                      \\
                                    & VSGNet\citep{ulutan2020vsgnet}     &84.9 & ResNet-152                & 19.80          & 16.05          & 20.91          & -              & -              & -              & \textit{51.8}                           & \textit{57.0}                           \\
                                    & DRG\citep{gao2020drg}              &46.1 & ResNet50-FPN               &21.66    &{\ul19.66}            &22.25          & -          & -          & -          & 51.0                                 & -                                    \\
                                    & SCG\citep{zhang2021spatially}      &53.9 & ResNet50-FPN              & \textbf{29.26}  
                                           & \textbf{24.61}          & \textbf{30.65}          & \textbf{32.87}       & \textbf{27.89}              & \textbf{34.35}    & \textbf{54.2}                        & \textbf{60.9}    \\\hline
\multirow{1}{*}{Green Learning}          & EHOI (Ours)                      &\textbf{45.9} & ResNet50-FPN              & {\ul24.53}    & 19.26    & {\ul26.09}    & {\ul27.64}    & {\ul22.70}    & {\ul29.12}    & 50.8                                 & 56.3                                 \\ \hline
                    
\end{tabular}
}
\end{table}


\begin{table}[htbp]
\caption{Detection performance comparison in mAP (\%) between EHOI and three end-to-end trained one-stage methods (without finetuning object and action detectors individually) for the HICO-Det dataset, where the mAP results of HOTR, AS-Net, and QPIC are taken from Ma et al.~\citep{ma2023fgahoi}.}
\centering
\begin{tabular}{lllllll}
\multirow{2}{*}{Methods} & \multicolumn{3}{c}{Default}                     \\ \cline{2-4}
                         & Full            & Rare           & Non-Rare      \\ \hline
HOTR \citep{kim2021hotr}                     & 23.46          & 16.21          & 25.65         \\
AS-Net \citep{chen2021reformulating}         & 24.40          & \textbf{22.39} & 25.01         \\
QPIC \citep{tamura2021qpic}                  & 24.21          & 17.51          & \textbf{26.21}         \\ \hline
EHOI (ours)                                 & \textbf{24.53} & {\ul19.26}     & {\ul26.09}
\end{tabular}
 
\label{table: Frozen Detector}
\end{table}

\begin{table}[htbp]
\caption{The comparison of FLOP numbers per query between EHOI and two SOTA two-stage 
models.}\label{table: FLOPs}
\centering
\begin{tabular}{llll}
\multirow{2}{*}{Two-Stage Models} & Default & \multirow{2}{*}{Parameters} & FLOPs  \\ \cline{2-2}
                      & Full         &                         &  (per query)           \\ \hline
SCG  (Graph)          & 29.26        & 53.9M (1.2x)            & 54M (4,500x)            \\
UPT  (Transformer)    & 32.62        & 54.7M (1.2x)            & 190M (15,800x)          \\ \hline
EHOI (Ours)           & 24.53        & \textbf{45.9M (1x)}     & \textbf{12K (1x)}                 
\end{tabular}
\end{table}

\section{Experiments}\label{sec:exp}

\subsection{Datasets}

V-COCO~\citep{gupta2015visual} and HICO-DET~\citep{chao2018learning} are two commonly used HOI detection datasets. V-COCO is a subset of the MS-COCO dataset. It contains 2,533 training images, 2,867 validation images, 4,946 test images, and 24 actions. HICO-DET is larger than V-COCO. It comprises 37,633 training images, 9,546 test images, 117 actions, and 600 interactions for various action-object pairs. Its training set has 117,871 human–object pairs with annotated bounding boxes, while its testing set contains 33,405 such pairs. HICO-DET is a challenging dataset. The 600 labeled interactions can be divided into 138 rare cases and 462 non-rare cases. Rare cases have less than ten samples in the training set. We use the mean Average Precision (mAP) as the evaluation metric, which is the mean of the average precision of all classes. 

The HICO-DET dataset has two settings: Default and Known Object. In the Default setting, the algorithms must simultaneously label human and object bounding boxes, object type, and relationship. In the Known Object setting, the algorithms already have object information and only need to label human and object bounding boxes and their relationship.

In the V-COCO setting, there are two scenarios: S1 and S2. Scenario 1 (S1) applies to test cases with missing role annotations. An agent role prediction is correct if the action is accurate, the overlap between the person boxes is >0.5, and the corresponding role is empty. Scenario 2 (S2) also applies to test cases with missing role annotations. An agent role prediction is correct if the action is correct and the overlap between the person boxes is >0.5 (ignoring the corresponding role). This scenario is suitable for cases with roles outside the COCO categories.

\subsection{Parameter Settings}

We used the pre-trained first-stage object detector provided by SCG~\citep{zhang2021spatially}. The human and object representations are the ROI pooling features from the CNN backbones. The feature selection processes are set to choose the top 1000 discriminant features from the human and object queries. The focal loss in the feature selection module is set to $\alpha = 1.0$ and $\gamma = 2.0$.

The conditional classifiers are based on the HICO-DET dataset in the second stage. The 600 interaction triples can be divided into 80 objects and 117 relations. The possible relation classes range from 2 to 16 by giving the object class. Hence, the Hamming codes can be represented by four binary bits with three error correlation bits. The bit classifiers are the XGBoost classifiers~\citep{chen2016xgboost} with 700 estimators and a depth of 5. The aggregation processes are conducted by Linear Discriminant Analysis(LDA). The one-hot coding classifiers follow the same setting as the bit classifier. 

\subsection{Experimental Results}

\subsubsection{Performance Benchmarking against Other Two-stage Models}

We compare the performance of EHOI against other SOTA two-stage models, including multi-stream and graph-based models, in Table \ref{table: Two-stage}, where the top and the second performers are in bold and underlined, respectively. EHOI achieves the second-best mAP values in most categories for HICO-DET. Our EHOI model also has the smallest number of learnable parameters. It's important to note that EHOI relies solely on visual features and doesn't incorporate external word embeddings or pose estimation information during training.

DRG~\citep{gao2020drg} uses extra language models in the training and prediction, which increases the computational burden in the training stage. Besides, graph-based HOI models need iterations of operations in graph convolutional networks, leading to a higher computation complexity, which will be discussed at the end of this subsection. SCG~\citep{zhang2021spatially} provides a similar mAP performance, but the classifier is a black box. The hidden layers are trained through end-to-end optimizations. However, our EHOI can give a statistical reasoning process for every intermediate classifier. The decision process is feedforward and can provide a sensible decision-making process for human understanding.  
\subsubsection{Performance Benchmarking against One-stage Models} 

Transformer models achieve impressive performance in various computer vision tasks at the expense of high computational complexities in training and inference. In the encoder-decoder-based detector, the model requires auxiliary queries for detection. There is no rule of thumb to determine the hyperparameters of the queries and save on computation requirements. Besides model efficiency, the training process is nontrivial for transformer-based one-stage HOI models. 

Especially, HICO-DET is a dataset with a long-tailed distribution. The model will become biased if we enforce the model to fit the skewed label distribution. The number of rare cases with less than ten samples has more than $10^6$ labeled pairs. The performance of one-stage models would drop dramatically if no finetuning were conducted on object detection and relation detection individually. Compared to the object labels, the interaction labels are incomplete ~\ref{fig: examples}. Hence, the imbalance labels will make the interaction detector the biased objected detector.

To verify the biased one-stage detectors, we compare the performance of our EHOI method and those without delicate finetuning in Table \ref{table: Frozen Detector}. It was observed in \citep{zhu2023diagnosing} that imbalanced triples tend to decrease the object detection performance. In contrast, our two-stage EHOI method does not need iterative finetuning on object and interaction detection. Thus, our architecture is more robust in terms of imbalanced triples. It can retain the performance of object detectors and plug-on relation detection features. 

\subsubsection{Comparison of Computational Complexity and Carbon Footprint} 

One of the critical aspects of our approach is the computational cost, with Green Learning explicitly focusing on energy consumption in the algorithms. Our claim of being "Green" is verified by comparing the computational cost of EHOI, UPT~\citep{zhang2022efficient}, and SCG~\citep{zhang2021spatially} in Table \ref{table: FLOPs}. UPT is a transformer-based method, while SCG is a graph-based method. They are SOTA two-stage HOI methods (see Figure \ref{fig: complexity} and Table \ref{table: Two-stage}).

Both of them demand a large number of iterated computations. The table shows that in the inference stage, the FLOP number per query of EHOI is 1/4,500 and 1/15,800 of that of SCG and UPT, respectively. Despite the possibility of accelerating tensor operations through parallel computing, the two deep learning models require much more electricity during the inference stage. EHOI is the most eco-friendly option in terms of carbon footprint. 

\subsection{Ablation Studies}

We conduct an ablation study to evaluate the individual module's functionality in section~\ref{ab: modules}. Table~\ref{table: ablation} compares the performance of our model under different settings. In section~\ref{ab: coding}, we provide the detailed difference in detection results between one-hot and Hamming coding with error correction codes. Figure~\ref{fig: mAP_diff} shows the advantage of the hybrid coding in the proposed HOI. The qualitative visual examples are illustrated in section~\ref{ab: visual}. Figure~\ref{fig: set_example} and ~\ref{fig: lose_example} visualize the effect of the two-query structure in the proposed EHOI detector.

In this section, all the detection experiments and the visualizations are conducted in the HICO-DET dataset under the default setting. We follow the result format in section~\ref{sec:exp} and use the mean Average Precision evaluation metric in the following subsections.

\subsubsection{Modular Design}\label{ab: modules}

Modular design is one of the highlights of our method. The feedforward components include feature selection, representation construction, and hybrid coding.


Two experiments are compared to study the feature selection effect: one using the raw features from the first-stage detector model and the other using the 1,000 most discriminative features selected through feature selection (i.e., Exp \#1 and 2 in Table ~\ref{table: ablation}). We observe that the performance improves when more discriminative and less noisy features are used as input.

Exp \#2 and 3 demonstrate the impact of decoupling in conditional probability by linear combination proposed equation~\ref{eq: decouple}. Without the decomposition, the estimator takes all the selected features from the human and object RoI representations as the input and predicts the possible relationship among the 117 classes. In contrast, the two-query structure comprises two estimators taking human and object features, respectively.

Exp \#3 and the proposed EHOI compare the conventional one-hot and the proposed hybrid coding schemes. The hybrid coding scheme can benefit all categories in the HICO-DET dataset. The detailed comparison between the coding scheme and the number of parameters is revealed in Table~\ref{table: coding}. The modularized experiments are the evidence for the decomposition process in section~\ref{sec: decompose}. The subproblem formulations can improve the estimators. In summary, the modules in the proposed EHOI are crucial to achieve the best results in HOI detection.

\begin{table}[htbp]\label{table: ablation}
\caption{The ablation studies of the single modules. Feature Selection refers to Module C in Figure~\ref{fig: arch}. Two-query is the feature construction process, which divides the features into two predictions. The hybrid coding scheme involves coding with binary codes and hamming error correction codes.}
\centering
\begin{tabular}{ccccccc}
\multirow{2}{*}{Experiments}&\multirow{2}{*}{Feature Selection}&\multirow{2}{*}{Two-query}&\multirow{2}{*}{Coding} & \multicolumn{3}{c}{Default}\\ \cline{5-7}
           &          &          &          & Full           & Rare           & Non-Rare      \\ \hline
Exp \#1    &          &          &          & 11.83          & 9.78           & 12.44         \\
Exp \#2    &\checkmark&          &          & 16.01          & 10.97          & 17.52         \\
Exp \#3    &\checkmark&\checkmark&          & 20.55          & 13.47          & 22.66         \\ \hline
EHOI (ours)&\checkmark&\checkmark&\checkmark& \textbf{24.53} & \textbf{19.26} & \textbf{26.09}
\end{tabular}
\end{table}

\subsubsection{Effect of Label Coding}\label{ab: coding}

\begin{figure*}[htbp]
\centering
\includegraphics[width=\linewidth]{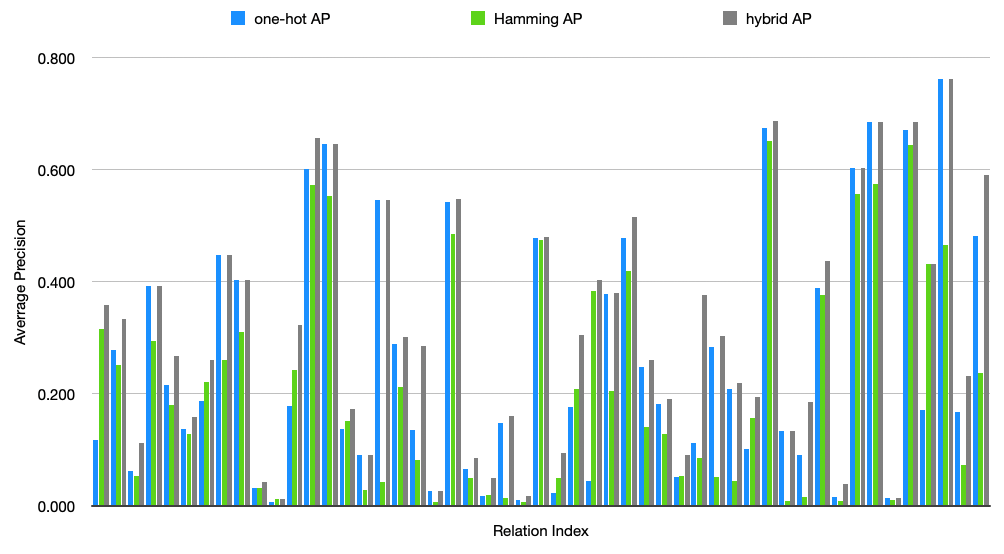}
\caption{The APs respect to different coding schemes. The one-hot coding in the interaction labels is represented by the blue histogram in the AP performance, while the green histogram depicts the Hamming coding performance. Additionally, the hybrid coding, which is proposed in our methodology, is illustrated by the gray histogram.} \label{fig: mAP_hist}
\end{figure*}

The relations between different coding schemes and detection performance are shown in Table.~\ref{table: coding}. We can further discuss precisions over the specific relation detection. We visualize the coding effects on the detection performance. One-hot and hamming coding can compensate for the performance in rare and non-rare cases. The results of one-hot coding, Hamming coding, and hybrid coding are illustrated in Figure~\ref{fig: mAP_hist}. As the proposed learning scheme, hybrid coding can provide the best detection performance. The two coding schemes can provide robustness and improve the detection performance by providing counterparts to each other.

In the detailed performance, one-hot coding and Hamming coding benefit different categories. Hamming coding can reach comparable performance when one-hot coding performs well. On the other hand, when the one-hot coding fails, the Hamming coding can provide the counterparts of the detection. 

\begin{figure*}[htbp]
\centering
\includegraphics[width=\linewidth]{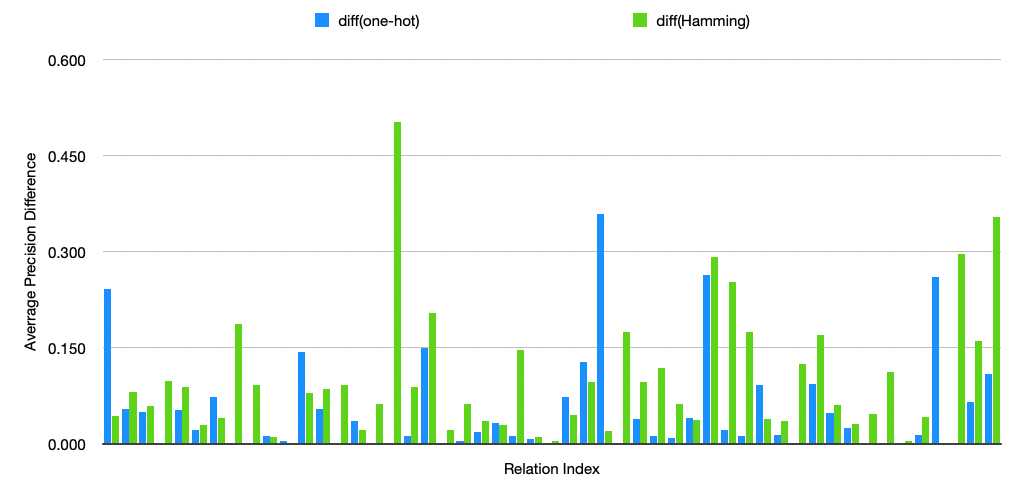}
\caption{The boost between the hybrid and on-hot coding and Hamming coding scheme. The difference between hybrid coding and one-hot coding is shown in blue. Similarly, the difference between hybrid and Hamming coding is shown in green.} \label{fig: mAP_diff}
\end{figure*}

Zooming on the difference between two coding schemes and hybrid coding, Figure~\ref{fig: mAP_diff} shows the performance increase along the relation categories. With the histogram, the performance does not result from using the two codes simultaneously. On the contrary, the relationship classes prefer a particular coding scheme. We can remove the extra bits using the performance boost as the criteria. Consequently, we can have the variable hybrid coding schemes for the corresponding relationship classes.

\subsubsection{Visual Explaination}\label{ab: visual}

To verify the decoupling in probability in Equation~\ref{eq: decouple}, we visualize the outputs from the human and object queries. In the example, we use the object type \textit{umbrella} and the corresponding relations \{\textit{carry}, \textit{hold}, \textit{lose}, \textit{no\_interaction}, \textit{open}, \textit{repair}, \textit{set}, \textit{stand\_under}\}. From human understanding, \{\textit{carry}, \textit{hold}, \textit{open}, \textit{stand\_under}\} may relate to the human representations. On the other hand, \{\textit{lose}, \textit{repair}, \textit{set}\} may relate to the object representations.

\begin{figure*}[htbp]
\centering
\includegraphics[width=\linewidth]{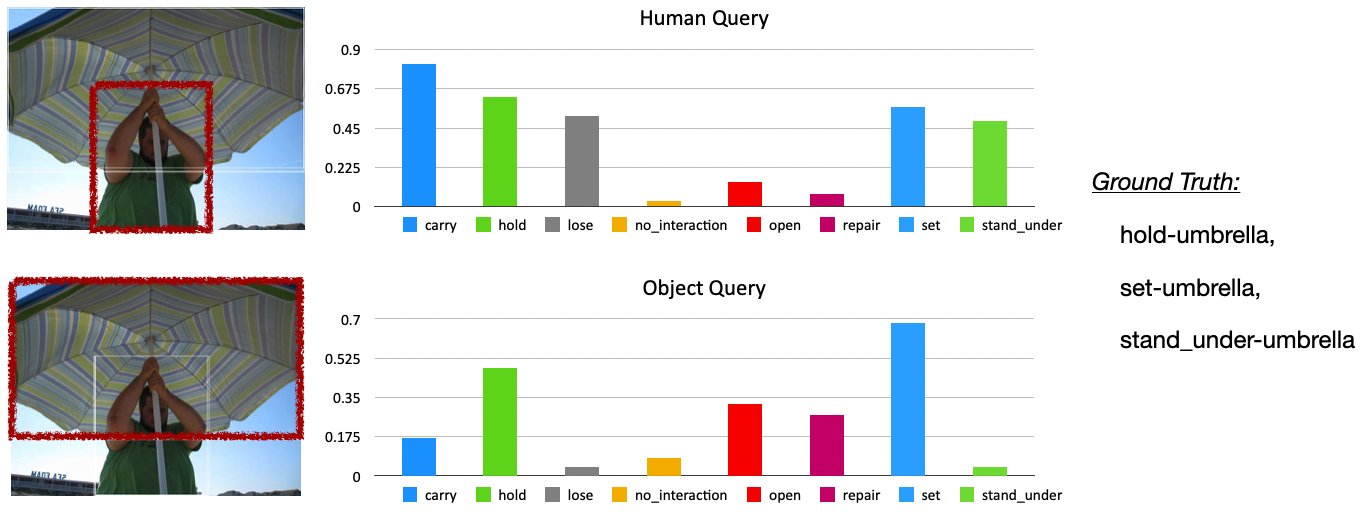}
\caption{The classifier in the human query can not find the critical representations in the umbrella region of interest. Fortunately, the object query can provide the counterparts of the prediction and make the outputs robust.} \label{fig: set_example}
\end{figure*}

Figure~\ref{fig: set_example} demonstrates the prediction from the human and object queries. The first row is the prediction from the human query, and the second is the prediction from the object query. The human query may confuse the predictions in the \textit{lose} and \textit{set} relations. In contrast, the object query can precisely capture the object status and correctly predict the \textit{lose} and \textit{set}.

\begin{figure*}[htbp]
\centering
\includegraphics[width=\linewidth]{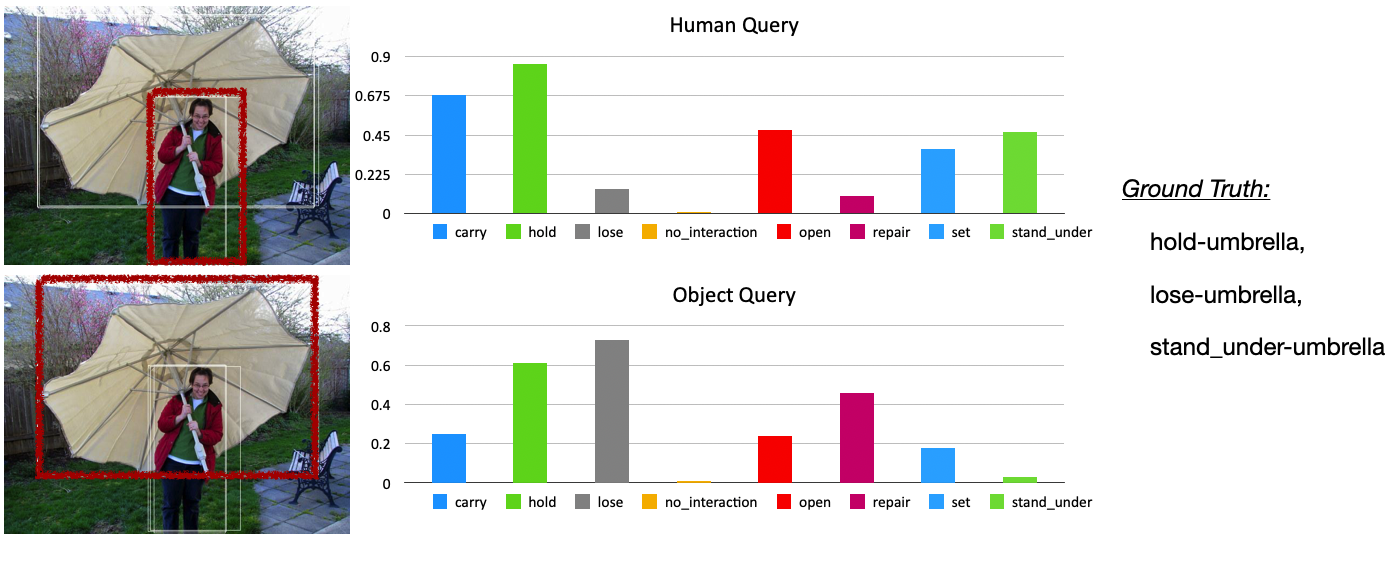}
\caption{The human representation is similar to Figure~\ref{fig: set_example}. The human query provides a probability distribution similar to the previous example. In contrast, the object query provides the \textit{lose} label.} \label{fig: lose_example}
\end{figure*}

Aiming to further understand the classifiers' behaviors, we give an example in Figure~\ref{fig: lose_example}. The human representations are similar in Figures~\ref{fig: set_example} and ~\ref{fig: lose_example}, which can be revealed in the ground truth labels \textit{hold} and \textit{stand\_under}. However, human queries can not detect the relations between the \textit{set} and \textit{lose}. The object queries can detect the two relations. In the object-query classifiers, the \textit{set} and \textit{lose} owe the highest probability scores in the results, leading to high confidence in the predictions. 

The decision process can be treated as human understanding. We can take the triple <human, relation, object> as two pairs: <human, relation> and <relation, object>. The human query is based on the information that describes the <human, relation> pairs. In contrast, the object query captures the visual features of the objects and expresses the <relation, object> pairs. The final outputs of the triple can be calculated as the weighted sum of the two predictors using Equation.~\ref{eq: decouple}.

\section{Conclusion and Future Work}\label{sec:conclusion}

An efficient HOI detector, called EHOI, was proposed in this work. It is both mathematically transparent and computationally efficient while offering competitive detection performance. The use of ECC for the coding of rare interaction types helps improve the robustness of EHOI. The divide-and-conquer feedforward design can reduce the complexity of the modules. Meanwhile, the intermediate results can provide physical meaning to the decision-making processes. We can extend the green learning scheme to image understanding applications and create human-sensible learning algorithms.

\section*{Acknowledgments}\label{sec:acknowledgments}

This work was supported by the Army Research Laboratory (ARL) under
agreement W911NF2020157. Computation for the work was supported by the
University of Southern California’s Center for Advanced Research
Computing (carc.usc.edu). 

\bibliographystyle{elsarticle-num-names} 
\bibliography{refs}

\begin{thebibliography}{38}
\expandafter\ifx\csname natexlab\endcsname\relax\def\natexlab#1{#1}\fi
\providecommand{\url}[1]{\texttt{#1}}
\providecommand{\href}[2]{#2}
\providecommand{\path}[1]{#1}
\providecommand{\DOIprefix}{doi:}
\providecommand{\ArXivprefix}{arXiv:}
\providecommand{\URLprefix}{URL: }
\providecommand{\Pubmedprefix}{pmid:}
\providecommand{\doi}[1]{\href{http://dx.doi.org/#1}{\path{#1}}}
\providecommand{\Pubmed}[1]{\href{pmid:#1}{\path{#1}}}
\providecommand{\bibinfo}[2]{#2}
\ifx\xfnm\relax \def\xfnm[#1]{\unskip,\space#1}\fi
\bibitem[{Chao et~al.(2018)Chao, Liu, Liu, Zeng, and Deng}]{chao2018learning}
\bibinfo{author}{Y.-W. Chao}, \bibinfo{author}{Y.~Liu}, \bibinfo{author}{X.~Liu}, \bibinfo{author}{H.~Zeng}, \bibinfo{author}{J.~Deng},
\newblock \bibinfo{title}{Learning to detect human-object interactions},
\newblock in: \bibinfo{booktitle}{2018 ieee winter conference on applications of computer vision (wacv)}, \bibinfo{organization}{IEEE}, \bibinfo{year}{2018}, pp. \bibinfo{pages}{381--389}.
\bibitem[{Gupta et~al.(2019)Gupta, Schwing, and Hoiem}]{gupta2019no}
\bibinfo{author}{T.~Gupta}, \bibinfo{author}{A.~Schwing}, \bibinfo{author}{D.~Hoiem},
\newblock \bibinfo{title}{No-frills human-object interaction detection: Factorization, layout encodings, and training techniques},
\newblock in: \bibinfo{booktitle}{Proceedings of the IEEE/CVF International Conference on Computer Vision}, \bibinfo{year}{2019}, pp. \bibinfo{pages}{9677--9685}.
\bibitem[{Zhong et~al.(2021)Zhong, Ding, Qu, and Tao}]{zhong2021polysemy}
\bibinfo{author}{X.~Zhong}, \bibinfo{author}{C.~Ding}, \bibinfo{author}{X.~Qu}, \bibinfo{author}{D.~Tao},
\newblock \bibinfo{title}{Polysemy deciphering network for robust human--object interaction detection},
\newblock \bibinfo{journal}{International Journal of Computer Vision} \bibinfo{volume}{129} (\bibinfo{year}{2021}) \bibinfo{pages}{1910--1929}.
\bibitem[{Lim et~al.(2023)Lim, Baskaran, Lim, Wong, See, and Tistarelli}]{lim2023ernet}
\bibinfo{author}{J.~Lim}, \bibinfo{author}{V.~M. Baskaran}, \bibinfo{author}{J.~M.-Y. Lim}, \bibinfo{author}{K.~Wong}, \bibinfo{author}{J.~See}, \bibinfo{author}{M.~Tistarelli},
\newblock \bibinfo{title}{Ernet: An efficient and reliable human-object interaction detection network},
\newblock \bibinfo{journal}{IEEE Transactions on Image Processing} \bibinfo{volume}{32} (\bibinfo{year}{2023}) \bibinfo{pages}{964--979}.
\bibitem[{Chen and Yanai(2023)}]{chen2023qahoi}
\bibinfo{author}{J.~Chen}, \bibinfo{author}{K.~Yanai},
\newblock \bibinfo{title}{Qahoi: Query-based anchors for human-object interaction detection},
\newblock in: \bibinfo{booktitle}{2023 18th International Conference on Machine Vision and Applications (MVA)}, \bibinfo{organization}{IEEE}, \bibinfo{year}{2023}, pp. \bibinfo{pages}{1--5}.
\bibitem[{Ma et~al.(2023)Ma, Wang, Wang, and Wei}]{ma2023fgahoi}
\bibinfo{author}{S.~Ma}, \bibinfo{author}{Y.~Wang}, \bibinfo{author}{S.~Wang}, \bibinfo{author}{Y.~Wei},
\newblock \bibinfo{title}{Fgahoi: Fine-grained anchors for human-object interaction detection},
\newblock \bibinfo{journal}{arXiv preprint arXiv:2301.04019}  (\bibinfo{year}{2023}).
\bibitem[{Kim et~al.(2021)Kim, Lee, Kang, Kim, and Kim}]{kim2021hotr}
\bibinfo{author}{B.~Kim}, \bibinfo{author}{J.~Lee}, \bibinfo{author}{J.~Kang}, \bibinfo{author}{E.-S. Kim}, \bibinfo{author}{H.~J. Kim},
\newblock \bibinfo{title}{Hotr: End-to-end human-object interaction detection with transformers},
\newblock in: \bibinfo{booktitle}{Proceedings of the IEEE/CVF Conference on Computer Vision and Pattern Recognition}, \bibinfo{year}{2021}, pp. \bibinfo{pages}{74--83}.
\bibitem[{Zhang et~al.(2022)Zhang, Campbell, and Gould}]{zhang2022efficient}
\bibinfo{author}{F.~Z. Zhang}, \bibinfo{author}{D.~Campbell}, \bibinfo{author}{S.~Gould},
\newblock \bibinfo{title}{Efficient two-stage detection of human-object interactions with a novel unary-pairwise transformer},
\newblock in: \bibinfo{booktitle}{Proceedings of the IEEE/CVF Conference on Computer Vision and Pattern Recognition}, \bibinfo{year}{2022}, pp. \bibinfo{pages}{20104--20112}.
\bibitem[{Zhang et~al.(2021)Zhang, Campbell, and Gould}]{zhang2021spatially}
\bibinfo{author}{F.~Z. Zhang}, \bibinfo{author}{D.~Campbell}, \bibinfo{author}{S.~Gould},
\newblock \bibinfo{title}{Spatially conditioned graphs for detecting human-object interactions},
\newblock in: \bibinfo{booktitle}{Proceedings of the IEEE/CVF International Conference on Computer Vision}, \bibinfo{year}{2021}, pp. \bibinfo{pages}{13319--13327}.
\bibitem[{Hou et~al.(2021)Hou, Yu, Qiao, Peng, and Tao}]{hou2021detecting}
\bibinfo{author}{Z.~Hou}, \bibinfo{author}{B.~Yu}, \bibinfo{author}{Y.~Qiao}, \bibinfo{author}{X.~Peng}, \bibinfo{author}{D.~Tao},
\newblock \bibinfo{title}{Detecting human-object interaction via fabricated compositional learning},
\newblock in: \bibinfo{booktitle}{Proceedings of the IEEE/CVF Conference on Computer Vision and Pattern Recognition}, \bibinfo{year}{2021}, pp. \bibinfo{pages}{14646--14655}.
\bibitem[{Wang et~al.(2020)Wang, Yang, Danelljan, Khan, Zhang, and Sun}]{wang2020learning}
\bibinfo{author}{T.~Wang}, \bibinfo{author}{T.~Yang}, \bibinfo{author}{M.~Danelljan}, \bibinfo{author}{F.~S. Khan}, \bibinfo{author}{X.~Zhang}, \bibinfo{author}{J.~Sun},
\newblock \bibinfo{title}{Learning human-object interaction detection using interaction points},
\newblock in: \bibinfo{booktitle}{Proceedings of the IEEE/CVF Conference on Computer Vision and Pattern Recognition}, \bibinfo{year}{2020}, pp. \bibinfo{pages}{4116--4125}.
\bibitem[{Liao et~al.(2020)Liao, Liu, Wang, Chen, Qian, and Feng}]{liao2020ppdm}
\bibinfo{author}{Y.~Liao}, \bibinfo{author}{S.~Liu}, \bibinfo{author}{F.~Wang}, \bibinfo{author}{Y.~Chen}, \bibinfo{author}{C.~Qian}, \bibinfo{author}{J.~Feng},
\newblock \bibinfo{title}{Ppdm: Parallel point detection and matching for real-time human-object interaction detection},
\newblock in: \bibinfo{booktitle}{Proceedings of the IEEE/CVF Conference on Computer Vision and Pattern Recognition}, \bibinfo{year}{2020}, pp. \bibinfo{pages}{482--490}.
\bibitem[{Kuo and Madni(2023)}]{kuo2023green}
\bibinfo{author}{C.-C.~J. Kuo}, \bibinfo{author}{A.~M. Madni},
\newblock \bibinfo{title}{Green learning: Introduction, examples and outlook},
\newblock \bibinfo{journal}{Journal of Visual Communication and Image Representation} \bibinfo{volume}{90} (\bibinfo{year}{2023}) \bibinfo{pages}{103685}.
\bibitem[{Ren et~al.(2015)Ren, He, Girshick, and Sun}]{ren2015faster}
\bibinfo{author}{S.~Ren}, \bibinfo{author}{K.~He}, \bibinfo{author}{R.~Girshick}, \bibinfo{author}{J.~Sun},
\newblock \bibinfo{title}{Faster r-cnn: Towards real-time object detection with region proposal networks},
\newblock \bibinfo{journal}{Advances in neural information processing systems} \bibinfo{volume}{28} (\bibinfo{year}{2015}).
\bibitem[{Carion et~al.(2020)Carion, Massa, Synnaeve, Usunier, Kirillov, and Zagoruyko}]{carion2020end}
\bibinfo{author}{N.~Carion}, \bibinfo{author}{F.~Massa}, \bibinfo{author}{G.~Synnaeve}, \bibinfo{author}{N.~Usunier}, \bibinfo{author}{A.~Kirillov}, \bibinfo{author}{S.~Zagoruyko},
\newblock \bibinfo{title}{End-to-end object detection with transformers},
\newblock in: \bibinfo{booktitle}{European conference on computer vision}, \bibinfo{organization}{Springer}, \bibinfo{year}{2020}, pp. \bibinfo{pages}{213--229}.
\bibitem[{Zhu et~al.(2020)Zhu, Su, Lu, Li, Wang, and Dai}]{zhu2020deformable}
\bibinfo{author}{X.~Zhu}, \bibinfo{author}{W.~Su}, \bibinfo{author}{L.~Lu}, \bibinfo{author}{B.~Li}, \bibinfo{author}{X.~Wang}, \bibinfo{author}{J.~Dai},
\newblock \bibinfo{title}{Deformable detr: Deformable transformers for end-to-end object detection},
\newblock \bibinfo{journal}{arXiv preprint arXiv:2010.04159}  (\bibinfo{year}{2020}).
\bibitem[{Jia et~al.(2023)Jia, Yuan, He, Wu, Yu, Lin, Sun, Zhang, and Hu}]{jia2023detrs}
\bibinfo{author}{D.~Jia}, \bibinfo{author}{Y.~Yuan}, \bibinfo{author}{H.~He}, \bibinfo{author}{X.~Wu}, \bibinfo{author}{H.~Yu}, \bibinfo{author}{W.~Lin}, \bibinfo{author}{L.~Sun}, \bibinfo{author}{C.~Zhang}, \bibinfo{author}{H.~Hu},
\newblock \bibinfo{title}{Detrs with hybrid matching},
\newblock in: \bibinfo{booktitle}{Proceedings of the IEEE/CVF conference on computer vision and pattern recognition}, \bibinfo{year}{2023}, pp. \bibinfo{pages}{19702--19712}.
\bibitem[{Chen et~al.(2021)Chen, Liao, Liu, Chen, Wang, and Qian}]{chen2021reformulating}
\bibinfo{author}{M.~Chen}, \bibinfo{author}{Y.~Liao}, \bibinfo{author}{S.~Liu}, \bibinfo{author}{Z.~Chen}, \bibinfo{author}{F.~Wang}, \bibinfo{author}{C.~Qian},
\newblock \bibinfo{title}{Reformulating hoi detection as adaptive set prediction},
\newblock in: \bibinfo{booktitle}{Proceedings of the IEEE/CVF Conference on Computer Vision and Pattern Recognition}, \bibinfo{year}{2021}, pp. \bibinfo{pages}{9004--9013}.
\bibitem[{Tamura et~al.(2021)Tamura, Ohashi, and Yoshinaga}]{tamura2021qpic}
\bibinfo{author}{M.~Tamura}, \bibinfo{author}{H.~Ohashi}, \bibinfo{author}{T.~Yoshinaga},
\newblock \bibinfo{title}{Qpic: Query-based pairwise human-object interaction detection with image-wide contextual information},
\newblock in: \bibinfo{booktitle}{Proceedings of the IEEE/CVF Conference on Computer Vision and Pattern Recognition}, \bibinfo{year}{2021}, pp. \bibinfo{pages}{10410--10419}.
\bibitem[{Kuhn(1955)}]{kuhn1955hungarian}
\bibinfo{author}{H.~W. Kuhn},
\newblock \bibinfo{title}{The hungarian method for the assignment problem},
\newblock \bibinfo{journal}{Naval research logistics quarterly} \bibinfo{volume}{2} (\bibinfo{year}{1955}) \bibinfo{pages}{83--97}.
\bibitem[{Liao et~al.(2022)Liao, Zhang, Lu, Wang, Li, and Liu}]{liao2022gen}
\bibinfo{author}{Y.~Liao}, \bibinfo{author}{A.~Zhang}, \bibinfo{author}{M.~Lu}, \bibinfo{author}{Y.~Wang}, \bibinfo{author}{X.~Li}, \bibinfo{author}{S.~Liu},
\newblock \bibinfo{title}{Gen-vlkt: Simplify association and enhance interaction understanding for hoi detection},
\newblock in: \bibinfo{booktitle}{Proceedings of the IEEE/CVF Conference on Computer Vision and Pattern Recognition}, \bibinfo{year}{2022}, pp. \bibinfo{pages}{20123--20132}.
\bibitem[{Radford et~al.(2021)Radford, Kim, Hallacy, Ramesh, Goh, Agarwal, Sastry, Askell, Mishkin, Clark et~al.}]{radford2021learning}
\bibinfo{author}{A.~Radford}, \bibinfo{author}{J.~W. Kim}, \bibinfo{author}{C.~Hallacy}, \bibinfo{author}{A.~Ramesh}, \bibinfo{author}{G.~Goh}, \bibinfo{author}{S.~Agarwal}, \bibinfo{author}{G.~Sastry}, \bibinfo{author}{A.~Askell}, \bibinfo{author}{P.~Mishkin}, \bibinfo{author}{J.~Clark}, et~al.,
\newblock \bibinfo{title}{Learning transferable visual models from natural language supervision},
\newblock in: \bibinfo{booktitle}{International conference on machine learning}, \bibinfo{organization}{PMLR}, \bibinfo{year}{2021}, pp. \bibinfo{pages}{8748--8763}.
\bibitem[{Dai et~al.(2017)Dai, Qi, Xiong, Li, Zhang, Hu, and Wei}]{dai2017deformable}
\bibinfo{author}{J.~Dai}, \bibinfo{author}{H.~Qi}, \bibinfo{author}{Y.~Xiong}, \bibinfo{author}{Y.~Li}, \bibinfo{author}{G.~Zhang}, \bibinfo{author}{H.~Hu}, \bibinfo{author}{Y.~Wei},
\newblock \bibinfo{title}{Deformable convolutional networks},
\newblock in: \bibinfo{booktitle}{Proceedings of the IEEE international conference on computer vision}, \bibinfo{year}{2017}, pp. \bibinfo{pages}{764--773}.
\bibitem[{Xia et~al.(2022)Xia, Pan, Song, Li, and Huang}]{xia2022vision}
\bibinfo{author}{Z.~Xia}, \bibinfo{author}{X.~Pan}, \bibinfo{author}{S.~Song}, \bibinfo{author}{L.~E. Li}, \bibinfo{author}{G.~Huang},
\newblock \bibinfo{title}{Vision transformer with deformable attention},
\newblock in: \bibinfo{booktitle}{Proceedings of the IEEE/CVF conference on computer vision and pattern recognition}, \bibinfo{year}{2022}, pp. \bibinfo{pages}{4794--4803}.
\bibitem[{Zhu et~al.(2023)Zhu, Xie, Xie, and Jiang}]{zhu2023diagnosing}
\bibinfo{author}{F.~Zhu}, \bibinfo{author}{Y.~Xie}, \bibinfo{author}{W.~Xie}, \bibinfo{author}{H.~Jiang},
\newblock \bibinfo{title}{Diagnosing human-object interaction detectors},
\newblock \bibinfo{journal}{arXiv preprint arXiv:2308.08529}  (\bibinfo{year}{2023}).
\bibitem[{Hou et~al.(2020)Hou, Peng, Qiao, and Tao}]{hou2020visual}
\bibinfo{author}{Z.~Hou}, \bibinfo{author}{X.~Peng}, \bibinfo{author}{Y.~Qiao}, \bibinfo{author}{D.~Tao},
\newblock \bibinfo{title}{Visual compositional learning for human-object interaction detection},
\newblock in: \bibinfo{booktitle}{Computer Vision--ECCV 2020: 16th European Conference, Glasgow, UK, August 23--28, 2020, Proceedings, Part XV 16}, \bibinfo{organization}{Springer}, \bibinfo{year}{2020}, pp. \bibinfo{pages}{584--600}.
\bibitem[{Kato et~al.(2018)Kato, Li, and Gupta}]{kato2018compositional}
\bibinfo{author}{K.~Kato}, \bibinfo{author}{Y.~Li}, \bibinfo{author}{A.~Gupta},
\newblock \bibinfo{title}{Compositional learning for human object interaction},
\newblock in: \bibinfo{booktitle}{Proceedings of the European Conference on Computer Vision (ECCV)}, \bibinfo{year}{2018}, pp. \bibinfo{pages}{234--251}.
\bibitem[{Gao et~al.(2020)Gao, Xu, Zou, and Huang}]{gao2020drg}
\bibinfo{author}{C.~Gao}, \bibinfo{author}{J.~Xu}, \bibinfo{author}{Y.~Zou}, \bibinfo{author}{J.-B. Huang},
\newblock \bibinfo{title}{Drg: Dual relation graph for human-object interaction detection},
\newblock in: \bibinfo{booktitle}{Computer Vision--ECCV 2020: 16th European Conference, Glasgow, UK, August 23--28, 2020, Proceedings, Part XII 16}, \bibinfo{organization}{Springer}, \bibinfo{year}{2020}, pp. \bibinfo{pages}{696--712}.
\bibitem[{Girshick(2015)}]{girshick2015fast}
\bibinfo{author}{R.~Girshick},
\newblock \bibinfo{title}{Fast r-cnn},
\newblock in: \bibinfo{booktitle}{Proceedings of the IEEE international conference on computer vision}, \bibinfo{year}{2015}, pp. \bibinfo{pages}{1440--1448}.
\bibitem[{Hamming(1950)}]{hamming1950error}
\bibinfo{author}{R.~W. Hamming},
\newblock \bibinfo{title}{Error detecting and error correcting codes},
\newblock \bibinfo{journal}{The Bell system technical journal} \bibinfo{volume}{29} (\bibinfo{year}{1950}) \bibinfo{pages}{147--160}.
\bibitem[{Yang et~al.(2022)Yang, Wang, Fu, Kuo et~al.}]{yang2022supervised}
\bibinfo{author}{Y.~Yang}, \bibinfo{author}{W.~Wang}, \bibinfo{author}{H.~Fu}, \bibinfo{author}{C.-C.~J. Kuo}, et~al.,
\newblock \bibinfo{title}{On supervised feature selection from high dimensional feature spaces},
\newblock \bibinfo{journal}{APSIPA Transactions on Signal and Information Processing} \bibinfo{volume}{11} (\bibinfo{year}{2022}).
\bibitem[{Lin et~al.(2017)Lin, Goyal, Girshick, He, and Doll{\'a}r}]{lin2017focal}
\bibinfo{author}{T.-Y. Lin}, \bibinfo{author}{P.~Goyal}, \bibinfo{author}{R.~Girshick}, \bibinfo{author}{K.~He}, \bibinfo{author}{P.~Doll{\'a}r},
\newblock \bibinfo{title}{Focal loss for dense object detection},
\newblock in: \bibinfo{booktitle}{Proceedings of the IEEE international conference on computer vision}, \bibinfo{year}{2017}, pp. \bibinfo{pages}{2980--2988}.
\bibitem[{Chen and Guestrin(2016)}]{chen2016xgboost}
\bibinfo{author}{T.~Chen}, \bibinfo{author}{C.~Guestrin},
\newblock \bibinfo{title}{Xgboost: A scalable tree boosting system},
\newblock in: \bibinfo{booktitle}{Proceedings of the 22nd acm sigkdd international conference on knowledge discovery and data mining}, \bibinfo{year}{2016}, pp. \bibinfo{pages}{785--794}.
\bibitem[{Wan et~al.(2019)Wan, Zhou, Liu, Li, and He}]{wan2019pose}
\bibinfo{author}{B.~Wan}, \bibinfo{author}{D.~Zhou}, \bibinfo{author}{Y.~Liu}, \bibinfo{author}{R.~Li}, \bibinfo{author}{X.~He},
\newblock \bibinfo{title}{Pose-aware multi-level feature network for human object interaction detection},
\newblock in: \bibinfo{booktitle}{Proceedings of the IEEE/CVF International Conference on Computer Vision}, \bibinfo{year}{2019}, pp. \bibinfo{pages}{9469--9478}.
\bibitem[{Bansal et~al.(2020)Bansal, Rambhatla, Shrivastava, and Chellappa}]{bansal2020detecting}
\bibinfo{author}{A.~Bansal}, \bibinfo{author}{S.~S. Rambhatla}, \bibinfo{author}{A.~Shrivastava}, \bibinfo{author}{R.~Chellappa},
\newblock \bibinfo{title}{Detecting human-object interactions via functional generalization},
\newblock in: \bibinfo{booktitle}{Proceedings of the AAAI Conference on Artificial Intelligence}, volume~\bibinfo{volume}{34}, \bibinfo{year}{2020}, pp. \bibinfo{pages}{10460--10469}.
\bibitem[{Zhou and Chi(2019)}]{zhou2019relation}
\bibinfo{author}{P.~Zhou}, \bibinfo{author}{M.~Chi},
\newblock \bibinfo{title}{Relation parsing neural network for human-object interaction detection},
\newblock in: \bibinfo{booktitle}{Proceedings of the IEEE/CVF International Conference on Computer Vision}, \bibinfo{year}{2019}, pp. \bibinfo{pages}{843--851}.
\bibitem[{Ulutan et~al.(2020)Ulutan, Iftekhar, and Manjunath}]{ulutan2020vsgnet}
\bibinfo{author}{O.~Ulutan}, \bibinfo{author}{A.~Iftekhar}, \bibinfo{author}{B.~S. Manjunath},
\newblock \bibinfo{title}{Vsgnet: Spatial attention network for detecting human object interactions using graph convolutions},
\newblock in: \bibinfo{booktitle}{Proceedings of the IEEE/CVF conference on computer vision and pattern recognition}, \bibinfo{year}{2020}, pp. \bibinfo{pages}{13617--13626}.
\bibitem[{Gupta and Malik(2015)}]{gupta2015visual}
\bibinfo{author}{S.~Gupta}, \bibinfo{author}{J.~Malik},
\newblock \bibinfo{title}{Visual semantic role labeling},
\newblock \bibinfo{journal}{arXiv preprint arXiv:1505.04474}  (\bibinfo{year}{2015}).

\end{thebibliography}



\end{document}